\documentclass[letterpaper]{article} 
\usepackage{aaai2026}  
\usepackage{times}  
\usepackage{helvet}  
\usepackage{courier}  
\usepackage[hyphens]{url}  
\usepackage{graphicx} 
\usepackage{natbib}  
\usepackage{caption} 
\usepackage{placeins} 

\usepackage{algorithm}
\usepackage{algorithmic}

\usepackage{amsmath}
\usepackage{amsfonts} 

\usepackage{xcolor}
\definecolor{darkerGreen}{rgb}{0.0,0.7,0.0}
\definecolor{darkerRed}{rgb}{0.9,0.0,0.0}
\usepackage{booktabs}

\def\mA{{\mathbf{A}}}

\def\mD{{\mathbf{D}}}

\def\mH{{\mathbf{H}}}
\def\mI{{\mathbf{I}}}

\def\mS{{\mathbf{S}}}

\def\mW{{\mathbf{W}}}
\def\mX{{\mathbf{X}}}

\def\mZ{{\mathbf{Z}}}

\def\vb{{\mathbf{b}}}

\def\vx{{\mathbf{x}}}

\def\vz{{\mathbf{z}}}

\def\gC{{\mathcal{C}}}

\def\gE{{\mathcal{E}}}

\def\gG{{\mathcal{G}}}

\def\gV{{\mathcal{V}}}

\usepackage{newfloat}
\usepackage{listings}
\DeclareCaptionStyle{ruled}{labelfont=normalfont,labelsep=colon,strut=off} 
\floatstyle{ruled}
\newfloat{listing}{tb}{lst}{}
\floatname{listing}{Listing}
\title{Mini-Batch Class Composition Bias in Link Prediction\thanks{Accepted at GCLR 2026: the 5th Workshop on Graphs and more Complex Structures For Learning and Reasoning, colocated with AAAI 2026.}}

\author{
    Kieran Maguire,
    Srinandan Dasmahapatra
}

\affiliations{
    University of Southampton\\
    School of Electronics and Computer Science\\
    \{K.A.Maguire, sd\}@soton.ac.uk,
}

\begin{document}

\maketitle

\begin{abstract}
    Prior work on node classification has shown that Graph Neural Networks (GNNs) can learn representations that transfer across graphs, when underlying graph properties are shared. For a fixed graph, one would then expect GNNs trained for link prediction to learn a representation consistent with that learnt for node classification. We show this intuition does not hold in the general case. Instead, we find popular link prediction models can learn a trivial mini-batch dependent heuristic, enabled by batch-normalisation layers, to solve the edge classification task. When correcting for this, we observe increased alignment of the network representation with node-class relevant features, suggesting the network has learnt a graph representation that better aligns with the underlying graph's properties. Our findings suggest that standard link prediction training may be leading us to overestimate link predictors' ability to learn a generalised representation of a graph that is consistent across tasks.
\end{abstract}



\section{Introduction}

Equivariant node representations in graphs are beneficial in node classification, but the same is not true in link prediction ~\citep{srinivasan2019equivalence}. This contrast is one reason why distinct Graph Neural Network (GNN) architectures have emerged for each task. While separate tasks may require separate GNN design, prior works have shown that learnt representations of a graph for the same task can be transferred between different graphs. ~\citet{spectransfer} demonstrated transferability between graphs with different dimensions and topologies, only requiring that ``both graphs discretise the same underlying space in some generic sense''. Given that the learnt representations within one task can be transferred across graphs, one may assume this would extend to the learnt representation of one graph across tasks. That is, the intuition that GNNs trained for link prediction should also learn features that are predictive of node classes, especially when we consider homophilic graphs having the property that nodes of the same node class connect.

The hypothesis that a link predictor should also learn node class information is shown by research into the neighbourhood patterns of nodes and their influence on node classification. \citet{ma2022is} showed that as long as nodes from the same node class share similar node class neighbourhood patterns, GNNs can still produce strong node classification performance, even in heterophilic graphs. \citet{luan2023when} considered not only intra-class separation by neighbourhood pattern, but also inter-class separation. \citet{zheng2024what} further established that feature and structural homophily, rather than the node class label alone, drives the distinguishability of nodes and downstream node classification performance. This is corroborated by the results of \citet{Qian2019QuantifyingTA}, who showed that the extent to which feature and structural graph properties align with node class label is influential in GNN node classification tasks. We believe these combined works empirically demonstrate the theory proposed by \citet{Xu2018HowPA}. For a node $v$, define a class-wise neighbourhood signature

\begin{equation}
\mathcal{N}_c(v)\;=\;
\square_{\text{agg}}\!\Bigl(
  \bigl\{\!\vx_u \;\big|\; u\in\gV,\; \operatorname{dist}(u,v)\le k\bigr\}
\Bigr),
\label{eq:nc}
\end{equation}

obtained by applying an aggregation operator $\square_{\text{agg}}$ (e.g.\ mean, sum) to the feature vectors of all nodes within the $k$-hop radius around~$v$. If the aggregation function over the graph is injective, then any two nodes with isomorphic k-hop rooted subtrees map to the same embedding. In the ideal limit, all nodes of the same class collapse to a single prototype $\mathcal{N}_c(v)\approx\boldsymbol{\mu}_c$ for every node $v$ whose label is $c$. We can observe that this graph property is not dependent on the node class label, rather the node class label is characterised by it. Thus, even in the absence of $c$ as in link prediction, ~\eqref{eq:nc} can still produce a feature set that would align with a node class label. \citet{zhu2024on} showed that feature-homophily and feature-heterophily require different optimisation methods, and thus there is some \(\mathcal{N}(v)\) that affects the link predictor's ability to reduce the Binary Cross Entropy loss (BCE) loss. Thus, we can consider that during the training of a link predictor, there exists some nodes whose k-hop neighbourhood signatures \(\mathcal{N}(v)\) which concentrate near a class prototype \(\boldsymbol{\mu}_c\) as a result of reducing the BCE loss. As a result, we would assume that GNNs trained for link prediction should also learn features that are aligned with node class.

In this work, we identify that this intuition appears to not be true in common link prediction models. In the typical training procedure for link prediction, mini-batches are constructed to be approximately half positive, half negative edges. We find this introduces a bias whereby the link predictor can use a batch normalisation layer to learn this mini-batch class distribution as a heuristic to predict edges. As a result, the model can avoid learning more complex node class features. This implies that GNNs for link prediction can effectively “cheat” when learning graph features, and therefore considers the question of whether we may be overestimating the performance of current link prediction methods to learn graph representations that capture the underlying properties of the graph. Additionally, learning such a batch heuristic may limit future efforts to transfer the learnt graph representation in the link prediction task, as performed in prior works in the node classification task ~\citep{zhao2025fullyinductive}. To address this bias, we randomise the fraction of positive and negative edges per mini-batch. We find this leads to decreased link prediction performance, but improved alignment with features relevant to the node classification task, establishing that link prediction models can also learn features that are more strongly aligned with node class in the right training regime.

\section{Model and Training Procedure}

In this section, we describe the architecture of the link-predictor models we consider, and their training regime. We consider an undirected graph $\gG=(\gV,\gE)$ with $|\gV|=N$ nodes and $c$ classes, and node-class labelling ground-truth function $y\colon\gV\to\{\gC_1,\dots,\gC_c\}$. $\mA\in\mathbb{R}^{N\times N}$ is the adjacency matrix of the graph and $\mD=\operatorname{diag}(\mA\,\mathbf{1}_N)$ the corresponding degree matrix. Given $\tilde\mA=\mA+\mI_N$ and $\tilde\mD=\operatorname{diag}(\tilde\mA\,\mathbf{1}_N)$, then $\widehat\mA=\tilde\mD^{-1/2}\,\tilde\mA\,\tilde\mD^{-1/2}$ is the degree-normalised adjacency matrix with self-loops. Let \(P=\{\{u,v\}\subseteq \gV : u\neq v\}\) be the set of all possible unordered edges on \(\gV\) which are not self loops, with \(E\subseteq P\) the set of proposed edges. Any \(e\in E\) may be written as \(e=\{v_s,v_t\}\) with \(v_s,v_t\in \gV\). We define an edge labelling ground truth function $\phi: P \to \{0,1\},$ such that
\begin{equation}
\phi(e) =
\begin{cases}
1 & \text{if } e \in \gE, \\
0 & \text{if } e \notin \gE.
\end{cases}
\label{eq:phi_indicator}
\end{equation}

The goal is to learn a parameterised GNN model $\psi_{\Theta}(\cdot)$ for link prediction which minimises the empirical risk given by
\begin{equation}
\min_{\Theta} \frac{1}{|E|} \sum_{i=1}^{|E|} \mathcal{L}\big(\psi_{\Theta}(e_i \mid \gG),\,\phi(e_i)\big).
\label{eq:general_loss}
\end{equation}

We choose $\mathcal{L}$ based on the BCE loss. Let the model output probabilities \(\rho_i=\psi_{\Theta}(e_i \mid \gG)\in(0,1)\), obtained by applying a sigmoid to the model logits. Then
\begin{equation}
\min_{\Theta}\; \frac{1}{|E|}\sum_{i=1}^{|E|} \Big(
 -\,\phi(e_i)\,\log \rho_i
 - \big(1-\phi(e_i)\big)\,\log\!\big(1-\rho_i\big)
\Big).
\label{eq:bce_loss}
\end{equation}

Consider each node is provided with a feature vector $\vx_v \in \mathbb{R}^d$. Let $\mX \in \mathbb{R}^{N \times d}$ denote the data matrix containing the feature vectors of all $N$ nodes. We wish to provide $E$ with feature vectors derived from $\mX$ for learning. Node embeddings are first computed for the whole graph using a GNN encoder. Formally, for a GNN encoder with $L$ layers, let $\mH^{(0)}=\mX$, and update
\begin{equation}
\label{eq:gnn_formulation}
    \mH^{(l)} = \zeta\bigl(\widehat{\mA}\,\mH^{(l-1)}\,\mW^{(l)}\bigr), \qquad l=1,\dots,L,
\end{equation}
where $\zeta$ is the ReLU function and $\mW^{(l)} \in \mathbb{R}^{d_{l-1} \times d_{l}}$ is the weight matrix at layer $l$. Let $\mZ=\mH^{(L)}\in\mathbb{R}^{N\times d_L}$. 

Let $B\subseteq E$ be a training mini-batch of edges, with $K=|B|$. To provide edge-wise features for learning, we gather the corresponding source and target node embeddings from $\mZ$ for the edges in $B$ to form $\mZ_s,\mZ_t\in\mathbb{R}^{K\times d_L}$, which we stack as $\mZ_{s;t}\in\mathbb{R}^{K\times 2\times d_L}$. Following this, $\mZ_{s;t}$ undergoes an operation to collapse the second dimension to produce edge representations; writing the result with source and target subscripts, we have
\begin{equation}
\mZ_1 = \mZ_s \odot \mZ_t,
\label{eq:z1}
\end{equation}
where $\odot$ denotes the element-wise (Hadamard) product. $\mZ_1$ is then passed to an MLP layer with batch norm $\operatorname{BN}$ and ReLU $\zeta$ to produce
\begin{equation}
\mZ_2 = \zeta\!\bigl(\operatorname{BN}(\mZ_1 \mW_1 + \vb_1)\bigr),
\label{eq:z2}
\end{equation}
where $\mW_1\in\mathbb{R}^{d_L\times h}$ and $\vb_1\in\mathbb{R}^{h}$. $\mZ_2$ can then be concatenated with structural features $\mS$ from a given structural feature encoder $f$ defined as
\begin{equation}
\mS = f(\gG, E),
\label{eq:struct_feats}
\end{equation}
with $\mS\in\mathbb{R}^{K\times d_S}$. These outputs are then mapped to logits by a final linear layer as
\begin{equation}
\boldsymbol{\ell} = [\mZ_2;\mS]\,\mW_2 + \vb_2,
\qquad \boldsymbol{\ell}\in\mathbb{R}^{K},
\label{eq:final_layer}
\end{equation}
where $\mW_2\in\mathbb{R}^{\,h+d_S}$, $\vb_2\in\mathbb{R}$, and $;$ indicates the concatenation operation. The corresponding probabilities used in \eqref{eq:bce_loss} are
\begin{equation}
\rho_i \;=\; \sigma\!\big(\ell_i\big), \qquad i=1,\dots,K,
\label{eq:sigmoid_probs}
\end{equation}
where $\sigma$ is the logistic sigmoid and $\ell_i$ is the $i$-th entry of $\boldsymbol{\ell}$. We define the number of positive edges in the training mini-batch as
\begin{equation}
K^+ = \big|\{e \in B \mid e \in \gE\}\big|,
\qquad
K^- = \big|\{e \in B \mid e \notin \gE\}\big|.
\label{eq:KposKneg}
\end{equation}
By construction we have
\begin{equation}
K^+ + K^- = |B| = K.
\label{eq:batch_sum}
\end{equation}
Typically in the link prediction task, $K^+ \approx K^-$ for all mini-batches $B$ constructed in training. This arises since negative edges are generated in equal proportion to positives, shuffled, and then sampled at random for each mini-batch.

\begin{figure*}[!t]
  \centering
\includegraphics[width=1.0\textwidth,height=0.36\textheight,keepaspectratio]{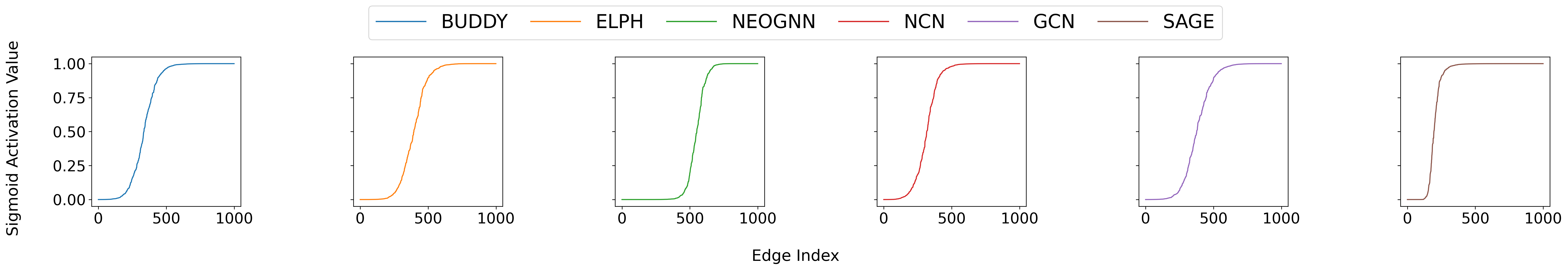}
  \caption{Sigmoid scores for a batch of all positive edges in the test set across multiple models in the Cora graph. A constant value of $1.0$ for all edges would indicate the model perfectly predicts all positive edges as correctly existing. Instead, we observe that approximately half the edges are predicted to not exist in most models, despite good model performance as measured by hits@k.}
  \label{fig:model_logits}
\end{figure*}

\section{Mini-batch Bias}
\label{sec:batch-bias}

We perform experiments with some of the most popular link predictors in terms of overall hits score and training speed, namely BUDDY \citep{chamberlain2023graph}, ELPH \citep{chamberlain2023graph}, NEOGNN \citep{yun2021neognns}, NCN \citep{wang2024neuralcommonneighborcompletion}, GCN \citep{kipf2017semisupervised}, and GraphSAGE \citep{hamilton2018inductiverepresentationlearninglarge}. We pass a batch of all positive training edges $e \in \gE$ through the trained link predictors and observe their logit distributions. To our surprise, we find that even in models with good link prediction performance as indicated by test Hits@K \citep{Bordes2013TranslatingEF}, each model predicts approximately half these edges as not existing in the batch, as seen in Figure~\ref{fig:model_logits}. One explanation for why this may be is that, at a fixed batch distribution, the batch norm operation can allow the model to learn a trivial batch-dependent heuristic. Assume that after some initial learning period, the network has learnt to output
\begin{equation}\label{eq:ai}
a_i =
\begin{cases}
1 & \text{if } \phi(e_i)=1,\\
0 & \text{if } \phi(e_i)=0,
\end{cases}
\end{equation}
regardless of the input features. Batch norm computes the batch mean
\begin{equation}\label{eq:mu_batch}
\mu \;=\; \frac{1}{K}\sum_{i=1}^K a_i \;=\; \frac{K^+}{K},
\end{equation}
and the batch variance
\begin{equation}\label{eq:sigma_batch}
\sigma^2 \;=\; \frac{1}{K}\sum_{i=1}^K \bigl(a_i - \mu\bigr)^2 \;=\; \mu(1-\mu) \;=\; \frac{K^+K^-}{K^2},
\end{equation}
so that \(\sigma=\sqrt{K^+K^-}/\,K\) is fixed once \(K^+\) and \(K^-\) are known. In the typical training procedure, \(K^+ \approx K^-\). Normalising each activation gives

\begin{equation}\label{eq:ahat_baseline_correct}
\hat a_i \;=\; \frac{a_i - \mu}{\sigma} \;=\;
\begin{cases}
\displaystyle +\sqrt{\frac{K^-}{K^+}} & \text{if } \phi(e_i)=1,\\[8pt]
\displaystyle -\sqrt{\frac{K^+}{K^-}} & \text{if } \phi(e_i)=0,
\end{cases}
\end{equation}

therefore all positives map to the same positive constant and all negatives to the same negative constant. Batch norm then applies a learnt affine transform
\begin{equation}\label{eq:zi}
z_i \;=\; \gamma\,\hat a_i + \beta.
\end{equation}
For any \(\gamma>0\) and suitable \(\beta\), this preserves the sign of \(\hat a_i\), so \(z_i>0\) for \(a_i=1\) and \(z_i<0\) for \(a_i=0\). In this way, batch normalisation first centres and scales the binary activations so that positives and negatives collapse to two constants; the subsequent affine transform \(\gamma\hat a_i+\beta\) only rescales and shifts these two values without changing their signs. This maintains a stable, monotone relationship between the activation and the logit, which can accelerate convergence; \(\beta\) adjusts the effective threshold without risking a reversal of the class encoding.

To prevent learning such a decision rule, consider \((K^+,K^-)=(p,K-p)\) with \(p \sim \mathcal{U}\{1,\dots,K-1\}\). Batch normalisation yields
\begin{equation}\label{eq:ahat_general}
\hat a_i \;=\; \frac{a_i - \mu}{\sigma} \;=\;
\begin{cases}
\displaystyle +\sqrt{\frac{K-p}{p}} & \text{if } \phi(e_i)=1,\\[10pt]
\displaystyle -\sqrt{\frac{p}{K-p}} & \text{if } \phi(e_i)=0,
\end{cases}
\end{equation}
which again yields one positive constant for all positive edges and one negative constant for all negative edges, but both depend on \(p\), the fraction of positive edges in the mini-batch. Applying the same affine transform \(z_i=\gamma \hat a_i + \beta\), the \(z_i\) now acquire an additional source of variability between batches via the randomness in \(p\). Since the choice of \(p\) is not correlated with edge class, any decision rule dependent on \(p\) will be destabilised, and thus should encourage the network to learn a more robust feature set (e.g., node-class–relevant features). As a result, we construct mini-batches with
\begin{equation}\label{eq:batch_construct}
K^+ = p,\qquad
K^- = K - p, \qquad
p \sim \mathcal{U}\{1,\dots,K-1\}.
\end{equation}
Explicitly, we randomly sample the edge class proportion of each mini-batch in a bootstrap procedure, so that the model cannot rely on edge class proportion as a heuristic. We refer to this mini-batching scheme as \textit{bias-corrected} mini-batching.

\section{Edge Separability by Node Class}
\label{section:Edge Separability by Node Class}

\begin{table*}[!t]
\centering
\scriptsize
\setlength{\tabcolsep}{3.5pt}
\renewcommand{\arraystretch}{1.0}
\caption{Comparison of Baseline, Bias-Corrected, and Change in hits@100 for BUDDY, ELPH, GCN, NCN, NEOGNN, and SAGE.}
\label{tab:LP_hits_all_models}

\resizebox{\textwidth}{!}{%
\begin{tabular}{l ccc ccc ccc}
\toprule
\textbf{Dataset} &
\multicolumn{3}{c}{\textbf{BUDDY}} &
\multicolumn{3}{c}{\textbf{ELPH}} &
\multicolumn{3}{c}{\textbf{GCN}} \\
& Original & Bias Correct & Change & Original & Bias Correct & Change & Original & Bias Correct & Change \\
\cmidrule(lr){2-4}\cmidrule(lr){5-7}\cmidrule(lr){8-10}
\midrule
\textbf{Cora}
& 85.291 \scriptsize{$\pm$ 1.859} & 84.344 \scriptsize{$\pm$ 2.415} & \textcolor{darkerRed}{-0.948}
& 85.726 \scriptsize{$\pm$ 1.349} & 85.561 \scriptsize{$\pm$ 1.761} & \textcolor{darkerRed}{-0.165}
& 83.909 \scriptsize{$\pm$ 1.326} & 83.455 \scriptsize{$\pm$ 1.502} & \textcolor{darkerRed}{-0.454} \\
\textbf{Citeseer}
& 92.218 \scriptsize{$\pm$ 1.242} & 92.027 \scriptsize{$\pm$ 1.031} & \textcolor{darkerRed}{-0.190}
& 91.442 \scriptsize{$\pm$ 1.251} & 90.871 \scriptsize{$\pm$ 1.329} & \textcolor{darkerRed}{-0.571}
& 90.095 \scriptsize{$\pm$ 1.387} & 88.884 \scriptsize{$\pm$ 1.345} & \textcolor{darkerRed}{-1.211} \\
\textbf{Pubmed}
& 74.524 \scriptsize{$\pm$ 2.383} & 73.141 \scriptsize{$\pm$ 2.576} & \textcolor{darkerRed}{-1.383}
& 68.876 \scriptsize{$\pm$ 1.623} & 68.987 \scriptsize{$\pm$ 1.890} & \textcolor{darkerGreen}{+0.111}
& 69.206 \scriptsize{$\pm$ 3.211} & 62.045 \scriptsize{$\pm$ 2.071} & \textcolor{darkerRed}{-7.160} \\
\textbf{CS}
& 72.136 \scriptsize{$\pm$ 1.387} & 75.978 \scriptsize{$\pm$ 1.350} & \textcolor{darkerGreen}{+3.841}
& 74.722 \scriptsize{$\pm$ 1.248} & 77.704 \scriptsize{$\pm$ 1.308} & \textcolor{darkerGreen}{+2.982}
& 73.859 \scriptsize{$\pm$ 1.470} & 75.107 \scriptsize{$\pm$ 1.127} & \textcolor{darkerGreen}{+1.248} \\
\textbf{Computers}
& 56.075 \scriptsize{$\pm$ 5.303} & 42.517 \scriptsize{$\pm$ 3.445} & \textcolor{darkerRed}{-13.558}
& 44.197 \scriptsize{$\pm$ 2.607} & 36.016 \scriptsize{$\pm$ 2.719} & \textcolor{darkerRed}{-8.181}
& 17.553 \scriptsize{$\pm$ 3.486} & 15.701 \scriptsize{$\pm$ 2.592} & \textcolor{darkerRed}{-1.852} \\
\textbf{Photo}
& 67.014 \scriptsize{$\pm$ 3.040} & 57.093 \scriptsize{$\pm$ 3.271} & \textcolor{darkerRed}{-9.921}
& 63.025 \scriptsize{$\pm$ 3.663} & 53.660 \scriptsize{$\pm$ 3.864} & \textcolor{darkerRed}{-9.365}
& 28.261 \scriptsize{$\pm$ 6.434} & 27.006 \scriptsize{$\pm$ 5.210} & \textcolor{darkerRed}{-1.255} \\
\textbf{ogbl-collab}
& 65.887 \scriptsize{$\pm$ 0.509} & 64.374 \scriptsize{$\pm$ 0.603} & \textcolor{darkerRed}{-1.513}
& 66.587 \scriptsize{$\pm$ 0.560} & 63.390 \scriptsize{$\pm$ 0.838} & \textcolor{darkerRed}{-3.197}
& 42.047 \scriptsize{$\pm$ 1.257} & 16.357 \scriptsize{$\pm$ 1.703} & \textcolor{darkerRed}{-25.689} \\
\textbf{ogbl-ppa}
& 48.717 \scriptsize{$\pm$ 0.902} & 37.349 \scriptsize{$\pm$ 2.642} & \textcolor{darkerRed}{-11.368}
& \multicolumn{3}{c}{OOM}
& 5.358 \scriptsize{$\pm$ 0.268} & 4.824 \scriptsize{$\pm$ 0.524} & \textcolor{darkerRed}{-0.534} \\
\bottomrule
\end{tabular}
}

\vspace{6pt}

\resizebox{\textwidth}{!}{%
\begin{tabular}{l ccc ccc ccc}
\toprule
\textbf{Dataset} &
\multicolumn{3}{c}{\textbf{NCN}} &
\multicolumn{3}{c}{\textbf{NEOGNN}} &
\multicolumn{3}{c}{\textbf{SAGE}} \\
& Original & Bias Correct & Change & Original & Bias Correct & Change & Original & Bias Correct & Change \\
\cmidrule(lr){2-4}\cmidrule(lr){5-7}\cmidrule(lr){8-10}
\midrule
\textbf{Cora}
& 84.975 \scriptsize{$\pm$ 1.292} & 84.600 \scriptsize{$\pm$ 1.556} & \textcolor{darkerRed}{-0.375}
& 83.011 \scriptsize{$\pm$ 2.056} & 83.258 \scriptsize{$\pm$ 2.349} & \textcolor{darkerGreen}{+0.247}
& 74.284 \scriptsize{$\pm$ 1.576} & 74.659 \scriptsize{$\pm$ 1.891} & \textcolor{darkerGreen}{+0.375} \\
\textbf{Citeseer}
& 92.204 \scriptsize{$\pm$ 0.923} & 91.850 \scriptsize{$\pm$ 1.063} & \textcolor{darkerRed}{-0.354}
& 89.469 \scriptsize{$\pm$ 2.018} & 89.224 \scriptsize{$\pm$ 1.626} & \textcolor{darkerRed}{-0.245}
& 73.061 \scriptsize{$\pm$ 1.843} & 71.932 \scriptsize{$\pm$ 2.467} & \textcolor{darkerRed}{-1.129} \\
\textbf{Pubmed}
& 71.375 \scriptsize{$\pm$ 1.708} & 69.603 \scriptsize{$\pm$ 2.471} & \textcolor{darkerRed}{-1.772}
& 65.487 \scriptsize{$\pm$ 2.086} & 66.635 \scriptsize{$\pm$ 1.913} & \textcolor{darkerGreen}{+1.147}
& 45.310 \scriptsize{$\pm$ 3.037} & 36.593 \scriptsize{$\pm$ 3.263} & \textcolor{darkerRed}{-8.717} \\
\textbf{CS}
& 75.993 \scriptsize{$\pm$ 0.475} & 82.260 \scriptsize{$\pm$ 0.722} & \textcolor{darkerGreen}{+6.266}
& 70.575 \scriptsize{$\pm$ 0.799} & 72.323 \scriptsize{$\pm$ 1.669} & \textcolor{darkerGreen}{+1.749}
& 64.772 \scriptsize{$\pm$ 1.900} & 67.576 \scriptsize{$\pm$ 1.638} & \textcolor{darkerGreen}{+2.804} \\
\textbf{Computers}
& 48.262 \scriptsize{$\pm$ 1.443} & 47.930 \scriptsize{$\pm$ 1.540} & \textcolor{darkerRed}{-0.331}
& 20.594 \scriptsize{$\pm$ 2.722} & 11.991 \scriptsize{$\pm$ 3.120} & \textcolor{darkerRed}{-8.603}
& 8.929 \scriptsize{$\pm$ 1.300} & 8.495 \scriptsize{$\pm$ 2.130} & \textcolor{darkerRed}{-0.434} \\
\textbf{Photo}
& 62.507 \scriptsize{$\pm$ 3.417} & 57.521 \scriptsize{$\pm$ 2.946} & \textcolor{darkerRed}{-4.985}
& 29.390 \scriptsize{$\pm$ 5.525} & 24.156 \scriptsize{$\pm$ 5.276} & \textcolor{darkerRed}{-5.234}
& 17.413 \scriptsize{$\pm$ 3.716} & 17.108 \scriptsize{$\pm$ 3.347} & \textcolor{darkerRed}{-0.305} \\
\textbf{ogbl-collab}
& 65.165 \scriptsize{$\pm$ 0.154} & 65.175 \scriptsize{$\pm$ 0.134} & \textcolor{darkerGreen}{+0.010}
& 56.468 \scriptsize{$\pm$ 1.222} & 60.785 \scriptsize{$\pm$ 12.666} & \textcolor{darkerGreen}{+4.317}
& 39.303 \scriptsize{$\pm$ 1.794} & 4.221 \scriptsize{$\pm$ 1.567} & \textcolor{darkerRed}{-35.082} \\
\textbf{ogbl-ppa}
& 37.241 \scriptsize{$\pm$ 0.721} & 37.283 \scriptsize{$\pm$ 1.075} & \textcolor{darkerGreen}{+0.042}
& \multicolumn{3}{c}{OOM}
& 5.937 \scriptsize{$\pm$ 1.024} & 3.947 \scriptsize{$\pm$ nan} & \textcolor{darkerRed}{-1.990} \\
\bottomrule
\end{tabular}
}
\end{table*}

\begin{table*}[!t]
\centering
\scriptsize
\setlength{\tabcolsep}{3.5pt}
\renewcommand{\arraystretch}{1.0}
\caption{Comparison of TR between original and bias-corrected penultimate layer activations.}
\label{tab:z2_tr_all_models_2x3}

\resizebox{\textwidth}{!}{%
\begin{tabular}{l ccc ccc ccc}
\toprule
\textbf{Dataset} &
\multicolumn{3}{c}{\textbf{BUDDY}} &
\multicolumn{3}{c}{\textbf{ELPH}} &
\multicolumn{3}{c}{\textbf{GCN}} \\
& Original & Bias Correct & Change & Original & Bias Correct & Change & Original & Bias Correct & Change \\
\cmidrule(lr){2-4}\cmidrule(lr){5-7}\cmidrule(lr){8-10}
\midrule
\textbf{Cora}
& 0.059 \scriptsize{$\pm$ 0.005} & 0.140 \scriptsize{$\pm$ 0.009} & \textcolor{darkerGreen}{+0.081}
& 0.044 \scriptsize{$\pm$ 0.005} & 0.125 \scriptsize{$\pm$ 0.007} & \textcolor{darkerGreen}{+0.080}
& 0.039 \scriptsize{$\pm$ 0.001} & 0.165 \scriptsize{$\pm$ 0.010} & \textcolor{darkerGreen}{+0.126} \\
\textbf{Citeseer}
& 0.076 \scriptsize{$\pm$ 0.013} & 0.252 \scriptsize{$\pm$ 0.020} & \textcolor{darkerGreen}{+0.176}
& 0.059 \scriptsize{$\pm$ 0.007} & 0.099 \scriptsize{$\pm$ 0.008} & \textcolor{darkerGreen}{+0.041}
& 0.107 \scriptsize{$\pm$ 0.014} & 0.144 \scriptsize{$\pm$ 0.004} & \textcolor{darkerGreen}{+0.036} \\
\textbf{Pubmed}
& 0.013 \scriptsize{$\pm$ 0.001} & 0.238 \scriptsize{$\pm$ 0.008} & \textcolor{darkerGreen}{+0.225}
& 0.100 \scriptsize{$\pm$ 0.003} & 0.169 \scriptsize{$\pm$ 0.003} & \textcolor{darkerGreen}{+0.069}
& 0.022 \scriptsize{$\pm$ 0.001} & 0.239 \scriptsize{$\pm$ 0.006} & \textcolor{darkerGreen}{+0.217} \\
\textbf{CS}
& 0.225 \scriptsize{$\pm$ 0.002} & 0.728 \scriptsize{$\pm$ 0.005} & \textcolor{darkerGreen}{+0.503}
& 0.301 \scriptsize{$\pm$ 0.003} & 0.524 \scriptsize{$\pm$ 0.040} & \textcolor{darkerGreen}{+0.222}
& 0.179 \scriptsize{$\pm$ 0.002} & 0.864 \scriptsize{$\pm$ 0.005} & \textcolor{darkerGreen}{+0.686} \\
\textbf{Computers}
& 0.178 \scriptsize{$\pm$ 0.001} & 0.336 \scriptsize{$\pm$ 0.004} & \textcolor{darkerGreen}{+0.158}
& 0.289 \scriptsize{$\pm$ 0.003} & 0.454 \scriptsize{$\pm$ 0.003} & \textcolor{darkerGreen}{+0.164}
& 0.191 \scriptsize{$\pm$ 0.003} & 0.341 \scriptsize{$\pm$ 0.004} & \textcolor{darkerGreen}{+0.149} \\
\textbf{Photo}
& 0.233 \scriptsize{$\pm$ 0.003} & 0.646 \scriptsize{$\pm$ 0.011} & \textcolor{darkerGreen}{+0.413}
& 0.364 \scriptsize{$\pm$ 0.003} & 0.709 \scriptsize{$\pm$ 0.010} & \textcolor{darkerGreen}{+0.345}
& 0.251 \scriptsize{$\pm$ 0.003} & 0.700 \scriptsize{$\pm$ 0.028} & \textcolor{darkerGreen}{+0.449} \\
\bottomrule
\end{tabular}
}

\vspace{6pt}

\resizebox{\textwidth}{!}{%
\begin{tabular}{l ccc ccc ccc}
\toprule
\textbf{Dataset} &
\multicolumn{3}{c}{\textbf{NCN}} &
\multicolumn{3}{c}{\textbf{NEOGNN}} &
\multicolumn{3}{c}{\textbf{SAGE}} \\
& Original & Bias Correct & Change & Original & Bias Correct & Change & Original & Bias Correct & Change \\
\cmidrule(lr){2-4}\cmidrule(lr){5-7}\cmidrule(lr){8-10}
\midrule
\textbf{Cora}
& 0.056 \scriptsize{$\pm$ 0.004} & 0.131 \scriptsize{$\pm$ 0.007} & \textcolor{darkerGreen}{+0.075}
& 0.038 \scriptsize{$\pm$ 0.006} & 0.057 \scriptsize{$\pm$ 0.004} & \textcolor{darkerGreen}{+0.019}
& 0.035 \scriptsize{$\pm$ 0.002} & 0.176 \scriptsize{$\pm$ 0.012} & \textcolor{darkerGreen}{+0.141} \\
\textbf{Citeseer}
& 0.056 \scriptsize{$\pm$ 0.008} & 0.164 \scriptsize{$\pm$ 0.008} & \textcolor{darkerGreen}{+0.108}
& 0.060 \scriptsize{$\pm$ 0.019} & 0.085 \scriptsize{$\pm$ 0.014} & \textcolor{darkerGreen}{+0.025}
& 0.066 \scriptsize{$\pm$ 0.014} & 0.117 \scriptsize{$\pm$ 0.012} & \textcolor{darkerGreen}{+0.051} \\
\textbf{Pubmed}
& 0.075 \scriptsize{$\pm$ 0.001} & 0.169 \scriptsize{$\pm$ 0.006} & \textcolor{darkerGreen}{+0.095}
& 0.010 \scriptsize{$\pm$ 0.001} & 0.093 \scriptsize{$\pm$ 0.003} & \textcolor{darkerGreen}{+0.083}
& 0.012 \scriptsize{$\pm$ 0.000} & 0.018 \scriptsize{$\pm$ 0.001} & \textcolor{darkerGreen}{+0.006} \\
\textbf{CS}
& 0.197 \scriptsize{$\pm$ 0.002} & 0.246 \scriptsize{$\pm$ 0.014} & \textcolor{darkerGreen}{+0.049}
& 0.042 \scriptsize{$\pm$ 0.002} & 0.188 \scriptsize{$\pm$ 0.002} & \textcolor{darkerGreen}{+0.146}
& 0.111 \scriptsize{$\pm$ 0.004} & 0.631 \scriptsize{$\pm$ 0.064} & \textcolor{darkerGreen}{+0.520} \\
\textbf{Computers}
& 0.163 \scriptsize{$\pm$ 0.001} & 0.142 \scriptsize{$\pm$ 0.004} & \textcolor{darkerRed}{-0.021}
& 0.062 \scriptsize{$\pm$ 0.002} & 0.195 \scriptsize{$\pm$ 0.001} & \textcolor{darkerGreen}{+0.133}
& 0.039 \scriptsize{$\pm$ 0.000} & 0.649 \scriptsize{$\pm$ 0.008} & \textcolor{darkerGreen}{+0.610} \\
\textbf{Photo}
& 0.216 \scriptsize{$\pm$ 0.002} & 0.279 \scriptsize{$\pm$ 0.006} & \textcolor{darkerGreen}{+0.063}
& 0.157 \scriptsize{$\pm$ 0.005} & 0.367 \scriptsize{$\pm$ 0.005} & \textcolor{darkerGreen}{+0.210}
& 0.078 \scriptsize{$\pm$ 0.001} & 0.753 \scriptsize{$\pm$ 0.041} & \textcolor{darkerGreen}{+0.675} \\
\bottomrule
\end{tabular}
}
\end{table*}

\begin{table*}[!t]
\centering
\scriptsize
\setlength{\tabcolsep}{3.5pt}
\renewcommand{\arraystretch}{1.0}
\caption{Comparison of NMI across models and run types for BUDDY, ELPH, GCN, NCN, NEOGNN, and SAGE.}
\label{tab:NMI_comparison}

\resizebox{\textwidth}{!}{%
\begin{tabular}{l ccc ccc ccc}
\toprule
\textbf{Dataset} &
\multicolumn{3}{c}{\textbf{BUDDY}} &
\multicolumn{3}{c}{\textbf{ELPH}} &
\multicolumn{3}{c}{\textbf{GCN}} \\
& Original & Bias Correct & Change & Original & Bias Correct & Change & Original & Bias Correct & Change \\
\cmidrule(lr){2-4}\cmidrule(lr){5-7}\cmidrule(lr){8-10}
\midrule
\textbf{Cora}
& 0.208 \scriptsize{$\pm$ 0.020} & 0.365 \scriptsize{$\pm$ 0.016} & \textcolor{darkerGreen}{+0.157}
& 0.193 \scriptsize{$\pm$ 0.019} & 0.414 \scriptsize{$\pm$ 0.036} & \textcolor{darkerGreen}{+0.221}
& 0.099 \scriptsize{$\pm$ 0.020} & 0.387 \scriptsize{$\pm$ 0.018} & \textcolor{darkerGreen}{+0.288} \\
\textbf{Citeseer}
& 0.107 \scriptsize{$\pm$ 0.010} & 0.218 \scriptsize{$\pm$ 0.012} & \textcolor{darkerGreen}{+0.111}
& 0.113 \scriptsize{$\pm$ 0.012} & 0.215 \scriptsize{$\pm$ 0.011} & \textcolor{darkerGreen}{+0.101}
& 0.146 \scriptsize{$\pm$ 0.010} & 0.229 \scriptsize{$\pm$ 0.015} & \textcolor{darkerGreen}{+0.082} \\
\textbf{Pubmed}
& 0.005 \scriptsize{$\pm$ 0.000} & 0.151 \scriptsize{$\pm$ 0.003} & \textcolor{darkerGreen}{+0.146}
& 0.127 \scriptsize{$\pm$ 0.000} & 0.150 \scriptsize{$\pm$ 0.000} & \textcolor{darkerGreen}{+0.022}
& 0.023 \scriptsize{$\pm$ 0.004} & 0.241 \scriptsize{$\pm$ 0.019} & \textcolor{darkerGreen}{+0.218} \\
\textbf{CS}
& 0.218 \scriptsize{$\pm$ 0.004} & 0.621 \scriptsize{$\pm$ 0.021} & \textcolor{darkerGreen}{+0.402}
& 0.306 \scriptsize{$\pm$ 0.003} & 0.656 \scriptsize{$\pm$ 0.022} & \textcolor{darkerGreen}{+0.350}
& 0.247 \scriptsize{$\pm$ 0.007} & 0.663 \scriptsize{$\pm$ 0.019} & \textcolor{darkerGreen}{+0.416} \\
\textbf{Computers}
& 0.273 \scriptsize{$\pm$ 0.016} & 0.339 \scriptsize{$\pm$ 0.010} & \textcolor{darkerGreen}{+0.067}
& 0.241 \scriptsize{$\pm$ 0.015} & 0.317 \scriptsize{$\pm$ 0.011} & \textcolor{darkerGreen}{+0.076}
& 0.228 \scriptsize{$\pm$ 0.021} & 0.274 \scriptsize{$\pm$ 0.003} & \textcolor{darkerGreen}{+0.046} \\
\textbf{Photo}
& 0.454 \scriptsize{$\pm$ 0.029} & 0.570 \scriptsize{$\pm$ 0.002} & \textcolor{darkerGreen}{+0.116}
& 0.414 \scriptsize{$\pm$ 0.032} & 0.547 \scriptsize{$\pm$ 0.011} & \textcolor{darkerGreen}{+0.133}
& 0.398 \scriptsize{$\pm$ 0.015} & 0.518 \scriptsize{$\pm$ 0.012} & \textcolor{darkerGreen}{+0.120} \\
\bottomrule
\end{tabular}
}

\vspace{6pt}

\resizebox{\textwidth}{!}{%
\begin{tabular}{l ccc ccc ccc}
\toprule
\textbf{Dataset} &
\multicolumn{3}{c}{\textbf{NCN}} &
\multicolumn{3}{c}{\textbf{NEOGNN}} &
\multicolumn{3}{c}{\textbf{SAGE}} \\
& Original & Bias Correct & Change & Original & Bias Correct & Change & Original & Bias Correct & Change \\
\cmidrule(lr){2-4}\cmidrule(lr){5-7}\cmidrule(lr){8-10}
\midrule
\textbf{Cora}
& 0.215 \scriptsize{$\pm$ 0.015} & 0.362 \scriptsize{$\pm$ 0.020} & \textcolor{darkerGreen}{+0.148}
& 0.014 \scriptsize{$\pm$ 0.001} & 0.241 \scriptsize{$\pm$ 0.013} & \textcolor{darkerGreen}{+0.227}
& 0.239 \scriptsize{$\pm$ 0.029} & 0.495 \scriptsize{$\pm$ 0.004} & \textcolor{darkerGreen}{+0.256} \\
\textbf{Citeseer}
& 0.121 \scriptsize{$\pm$ 0.010} & 0.222 \scriptsize{$\pm$ 0.021} & \textcolor{darkerGreen}{+0.101}
& 0.018 \scriptsize{$\pm$ 0.002} & 0.147 \scriptsize{$\pm$ 0.013} & \textcolor{darkerGreen}{+0.129}
& 0.222 \scriptsize{$\pm$ 0.015} & 0.284 \scriptsize{$\pm$ 0.007} & \textcolor{darkerGreen}{+0.062} \\
\textbf{Pubmed}
& 0.119 \scriptsize{$\pm$ 0.001} & 0.094 \scriptsize{$\pm$ 0.003} & \textcolor{darkerRed}{-0.025}
& 0.003 \scriptsize{$\pm$ 0.000} & 0.100 \scriptsize{$\pm$ 0.004} & \textcolor{darkerGreen}{+0.098}
& 0.015 \scriptsize{$\pm$ 0.000} & 0.260 \scriptsize{$\pm$ 0.028} & \textcolor{darkerGreen}{+0.246} \\
\textbf{CS}
& 0.200 \scriptsize{$\pm$ 0.013} & 0.563 \scriptsize{$\pm$ 0.036} & \textcolor{darkerGreen}{+0.363}
& 0.019 \scriptsize{$\pm$ 0.000} & 0.342 \scriptsize{$\pm$ 0.017} & \textcolor{darkerGreen}{+0.323}
& 0.141 \scriptsize{$\pm$ 0.005} & 0.705 \scriptsize{$\pm$ 0.014} & \textcolor{darkerGreen}{+0.564} \\
\textbf{Computers}
& 0.232 \scriptsize{$\pm$ 0.002} & 0.329 \scriptsize{$\pm$ 0.021} & \textcolor{darkerGreen}{+0.096}
& 0.120 \scriptsize{$\pm$ 0.026} & 0.248 \scriptsize{$\pm$ 0.021} & \textcolor{darkerGreen}{+0.128}
& 0.060 \scriptsize{$\pm$ 0.002} & 0.351 \scriptsize{$\pm$ 0.006} & \textcolor{darkerGreen}{+0.291} \\
\textbf{Photo}
& 0.391 \scriptsize{$\pm$ 0.028} & 0.469 \scriptsize{$\pm$ 0.012} & \textcolor{darkerGreen}{+0.077}
& 0.217 \scriptsize{$\pm$ 0.003} & 0.436 \scriptsize{$\pm$ 0.036} & \textcolor{darkerGreen}{+0.219}
& 0.235 \scriptsize{$\pm$ 0.030} & 0.667 \scriptsize{$\pm$ 0.022} & \textcolor{darkerGreen}{+0.432} \\
\bottomrule
\end{tabular}
}
\end{table*}

\begin{table}[h!]
\centering
\caption{Comparison of TR between original and bias-corrected penultimate layer activations when ablating batch normalisation layers.}
\label{tab:norm_tr_ablation}
\resizebox{0.45\textwidth}{!}{%
\begin{tabular}{|l|ccc|ccc|}
\toprule
Dataset
  & \multicolumn{3}{c|}{\textbf{BUDDY}}
  & \multicolumn{3}{c|}{\textbf{GCN}} \\
  & Original & Bias Correct & Change
  & Original & Bias Correct & Change \\
\midrule
\textbf{Cora}
  & 0.035 \scriptsize{$\pm$ 0.002}
  & 0.034 \scriptsize{$\pm$ 0.002}
  & \textcolor{darkerRed}{-0.001}
  & 0.064 \scriptsize{$\pm$ 0.001}
  & 0.066 \scriptsize{$\pm$ 0.001}
  & \textcolor{darkerGreen}{+0.002} \\

\textbf{Citeseer}
  & 0.030 \scriptsize{$\pm$ 0.001}
  & 0.029 \scriptsize{$\pm$ 0.001}
  & \textcolor{darkerRed}{-0.001}
  & 0.041 \scriptsize{$\pm$ 0.002}
  & 0.041 \scriptsize{$\pm$ 0.002}
  & \textcolor{gray}{0.000} \\

\textbf{Pubmed}
  & 0.028 \scriptsize{$\pm$ 0.001}
  & 0.026 \scriptsize{$\pm$ 0.001}
  & \textcolor{darkerRed}{-0.002}
  & 0.024 \scriptsize{$\pm$ 0.000}
  & 0.026 \scriptsize{$\pm$ 0.000}
  & \textcolor{darkerGreen}{+0.002} \\
\bottomrule
\end{tabular}%
}
\end{table}

\begin{table}[h!]
\centering
\caption{Comparison of hits@$100$ between original and bias-corrected penultimate layer activations when ablating batch normalisation layers.}
\label{tab:norm_lp_hits}
\resizebox{0.45\textwidth}{!}{%
\begin{tabular}{|l|ccc|ccc|}
\toprule
Dataset & \multicolumn{3}{c|}{\textbf{BUDDY}} & \multicolumn{3}{c|}{\textbf{GCN}} \\
 & Original & Bias Correct & Change & Original & Bias Correct & Change \\
\midrule
\textbf{Cora}   
  & 64.294 \scriptsize{$\pm$ 1.074}   
  & 60.849 \scriptsize{$\pm$ 1.676}   
  & \textcolor{darkerRed}{-3.445}   
  & 71.594 \scriptsize{$\pm$ 1.574}   
  & 69.023 \scriptsize{$\pm$ 1.397}   
  & \textcolor{darkerRed}{-2.572} \\

\textbf{Citeseer} 
  & 66.912 \scriptsize{$\pm$ 1.576}   
  & 69.728 \scriptsize{$\pm$ 1.747}   
  & \textcolor{darkerGreen}{+2.816} 
  & 74.272 \scriptsize{$\pm$ 2.086}   
  & 75.007 \scriptsize{$\pm$ 2.329}   
  & \textcolor{darkerGreen}{+0.735} \\

\textbf{Pubmed}  
  & 61.665 \scriptsize{$\pm$ 1.889}   
  & 63.106 \scriptsize{$\pm$ 1.520}   
  & \textcolor{darkerGreen}{+1.441}   
  & 63.749 \scriptsize{$\pm$ 1.403}   
  & 63.066 \scriptsize{$\pm$ 1.375}   
  & \textcolor{darkerRed}{-0.683} \\
\bottomrule
\end{tabular}
}
\end{table}

\begin{figure}[!t]
  \centering
  \includegraphics[width=0.49\linewidth,keepaspectratio]{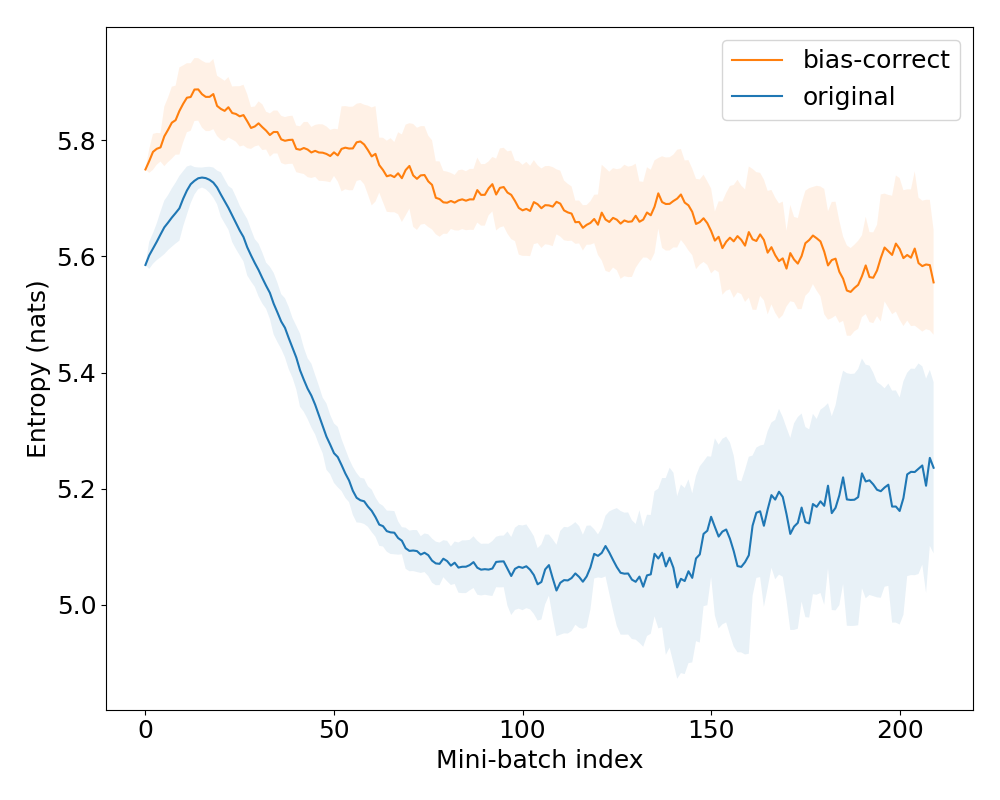}\hfill
  \includegraphics[width=0.49\linewidth,keepaspectratio]{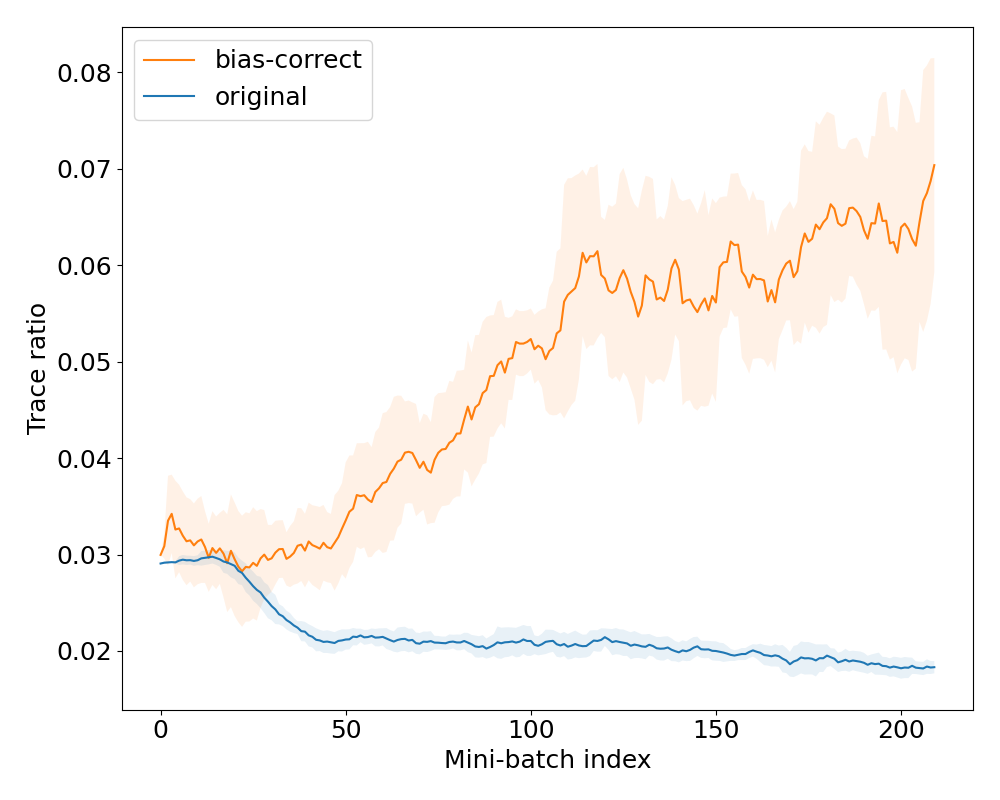}
  \caption{Effects of bias-corrected mini-batching in the gradients of $\mZ_2$ and the TR of $\mZ_2$ for each training mini-batch in BUDDY on Cora.}
  \label{fig:metrics_comparison}
\end{figure}

\begin{figure}[!t]
  \centering
  \includegraphics[width=1\linewidth]{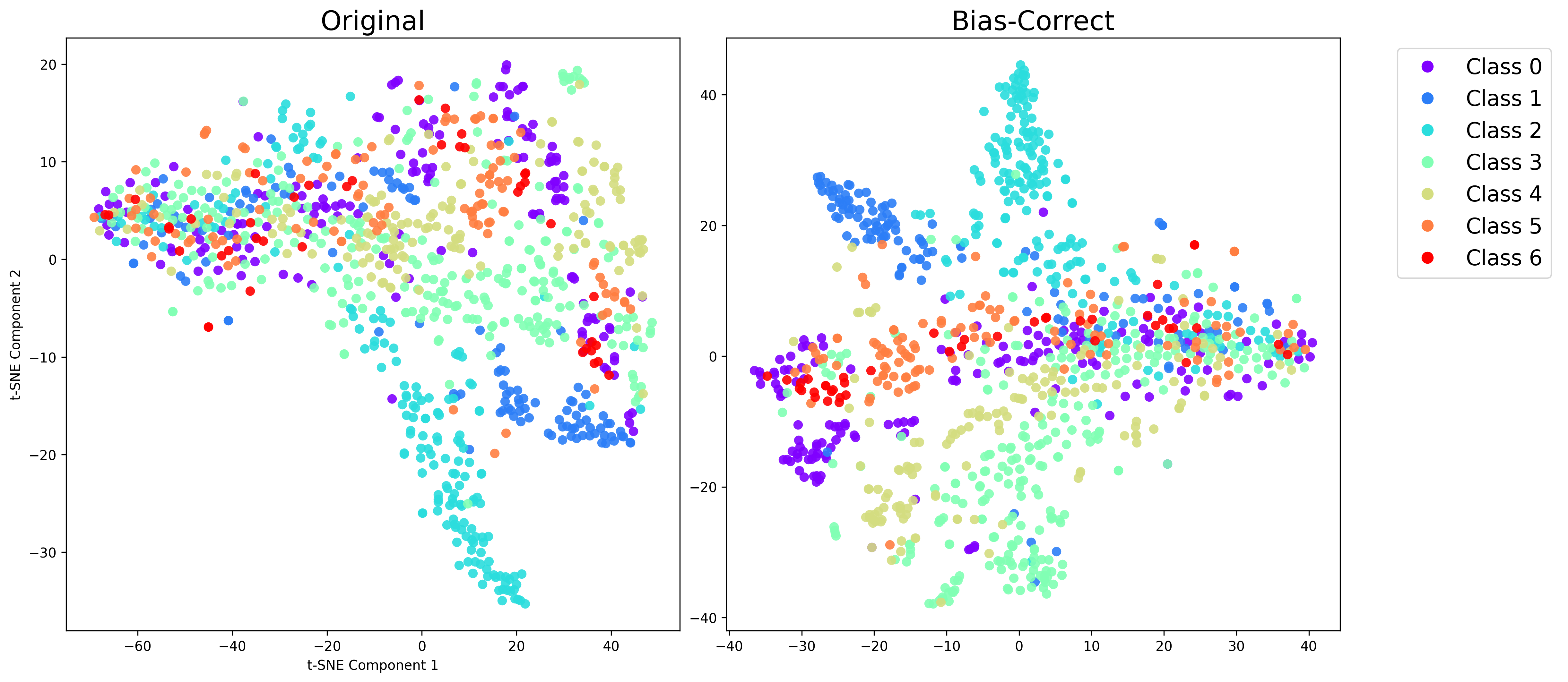}
  \caption{t-SNE \citep{Maaten2008VisualizingDU} of the standard training procedure (left) and bias-corrected (right) of $\mZ_2$ on Cora.}
  \label{fig:cora_tsne}
\end{figure}

To measure the extent to which node class relevant information is learnt, we label each edge by the node class label of the source node for each edge $y(v_s)$. We then use this label to find the Trace Ratio (TR) \citep{tirer2023perturbationanalysisneuralcollapse} of the edge embeddings, similar to \citet{kothapalli2023neuralcollapseperspectivefeature}, but reverse the denominator and numerator for readability. To fairly compare the extent to which the link predictors learn node class information across all models, we only consider the edge embeddings from the test set of edges in the graphs, at the penultimate layer of each network, $\mZ_2$ from \eqref{eq:z2}. Clustering by node class label $y(v_s)$ would indicate the networks find separating edges by features relevant to node class a necessary part of producing accurate edge predictions. Across all models, we use the same hyper-parameters provided by the model authors where possible. Each result is taken as the average over ten unique seeds, and standard deviations recorded. ``OOM'' indicates an out-of-memory error. 

Table \ref{tab:LP_hits_all_models} shows the change in hits@100 between the two mini-batching schemes. As expected, removing a bias that correlates strongly with edge class reduces link prediction performance on average. Table \ref{tab:z2_tr_all_models_2x3} shows as a result of our mini-batching change, we observe a consistently higher TR across all models, indicating models predictions are more dependent on features related to node class than before. This is visualised in Figure \ref{fig:cora_tsne}. Together, these results indicate that removing any learnable signal regarding batch composition has indeed led to increased learning of node class features and less dependency on batch class composition for forming predictions. Figure \ref{fig:metrics_comparison} shows the effects of our bias-corrected mini-batching in the gradients of $\mZ_2$ and the TR of $\mZ_2$ in each training mini-batch in BUDDY on Cora.

\section{Mapping to Node Features}
\label{section:K-means Clustering Performance}

We have observed that after correcting for bias in the mini-batching training regime, the link predictor does indeed appear to learn node class relevant information. In the spirit of ~\citet{spectransfer}, we also wish to observe how well this information transfers to the node classification task. This confirms in the most direct way the intuition that GNNs trained for link prediction also learn features that are predictive of node classes, and by extension if link and node tasks can share a common underlying representation. Recall that $\mZ_2$ is an edge embedding. If we are to perform a node level evaluation, the most direct way to do this is to translate this information back onto the node. We therefore take the average (mean) of all edge embeddings around each node in the graph, and assign this as the new node feature. Dropping the subscript from $\mZ_2$, this is defined as

\begin{equation}
\mathbf{x}_v' \;=\; \frac{1}{\lvert \mathcal{N}(v)\rvert}
\sum_{u \in \mathcal{N}(v)} \vz{_{u,v}},
\end{equation}

where $\mathcal{N}(v)$ denotes the set of 1-hop neighbours of node $v$ (i.e., all nodes $u$ for which $(u,v)$ is an edge), and $x_v'$ denotes the updated feature vector of node $v$. We then perform $k$-means clustering on these new node embeddings, with $k$ equal to the number of ground truth node classes. We measure the Normalised Mutual Information (NMI) \citep{Danon2005ComparingCS} to evaluate the global alignment of clusters and ground truth labels. Table \ref{tab:NMI_comparison},  provides further evidence that the link predictors indeed learn features relevant to the node classification task, provided any bias incurred in the training regime can be removed.

\section{Conclusion}
Our analysis shows that under the prevalent training practice of using mini-batches with an approximately fixed positive/negative edge fraction, link predictors can exploit a trivial mini-batch composition heuristic as a result of the batch normalisation layer(s) widely used across models. We find this inflates performance metrics at the expense of learning of node-class relevant features, and introduce a mini-batching scheme to randomise mini-batch class composition to account for this. Across multiple link prediction models, this change consistently increases the alignment of networks' internal representations with node class related features. When aggregated to nodes, we also find these node embeddings exhibit stronger agreement with node class labels under the bias-corrected mini-batching scheme. Our findings suggest standard link prediction evaluation can overestimate graph understanding if training permits batch-dependent heuristics, and adjustments to batching can steer models toward learning features which better align with the graphs structural properties without architectural change to the models themselves.

\bibliography{aaai2026}
\end{document}